\documentclass[10pt,twocolumn,letterpaper]{article}

\usepackage[pagenumbers]{cvpr} 

\usepackage{graphicx}
\usepackage{amsmath}
\usepackage{amssymb}
\usepackage{booktabs}

\usepackage[pagebackref,breaklinks,colorlinks]{hyperref}

\usepackage[capitalize]{cleveref}
\crefname{section}{Sec.}{Secs.}
\Crefname{section}{Section}{Sections}
\Crefname{table}{Table}{Tables}
\crefname{table}{Tab.}{Tabs.}

\def\cvprPaperID{14} 
\def\confName{CVPR}
\def\confYear{2022}

\begin{document}

\title{LANTERN-RD: Enabling Deep Learning for Mitigation of the Invasive Spotted Lanternfly}

\author{Srivatsa Kundurthy\\
The Academy for Mathematics, Science, and Engineering\\
{\tt\small kundurthys@acm.org}}

\maketitle

\begin{abstract}
  The Spotted Lanternfly (SLF) is an invasive planthopper that threatens the local biodiversity and agricultural economy of regions such as the Northeastern United States and Japan. As researchers scramble to study the insect, there is a great potential for computer vision tasks such as detection, pose estimation, and accurate identification to have important downstream implications in containing the SLF. However, there is currently no publicly available dataset for training such AI models. To enable computer vision applications and motivate advancements to challenge the invasive SLF problem, we propose LANTERN-RD, the first curated image dataset of the spotted lanternfly and its look-alikes, featuring images with varied lighting conditions, diverse backgrounds, and subjects in assorted poses. A VGG16-based baseline CNN validates the potential of this dataset for stimulating fresh computer vision applications to accelerate invasive SLF research. Additionally, we implement the trained model in a simple mobile classification application in order to directly empower responsible public mitigation efforts. The overarching mission of this work is to introduce a novel SLF image dataset and release a classification framework that enables computer vision applications, boosting studies surrounding the invasive SLF and assisting in minimizing its agricultural and economic damage. 
\end{abstract}


\section{Introduction}
\label{sec:intro}

\textit{Lycorma Delicatula}, dubbed the Spotted Lanternfly (SLF), has recently spread invasively from parts of Eastern Asia to several
other continents, establishing successfully in certain new environments due to reasons such as climate preference \cite{5, 6, 14} and the abundance of sustainable plant hosts \cite{1, 7, 8, 15}. Current regions of interest include the American Northeast and Japan, among others \cite{16}. As a new contender in the ecosystem and a nuisance pest, the SLF poses a significant threat to local agriculture and the economy through its destruction of fruit trees, ornamental trees, timber, vineyards, and building structures \cite{10, 11, 18}, potentially leading to billions of dollars of damage. Evolving studies propose several methods for containment, from biocontrol agents \cite{2, 3, 9} to trapping mechanisms \cite{4, 12, 17}. Notably, local environmental authorities have mandated quarantines to slow the spread of the SLF \cite{pennslf}, and are calling on the public to trap, report, and even take steps to exterminate sighted SLFs \cite{njslf2}. 

\begin{figure*}[htb]
    \centering 
\begin{subfigure}{0.25\textwidth}
  \includegraphics[width=\linewidth]{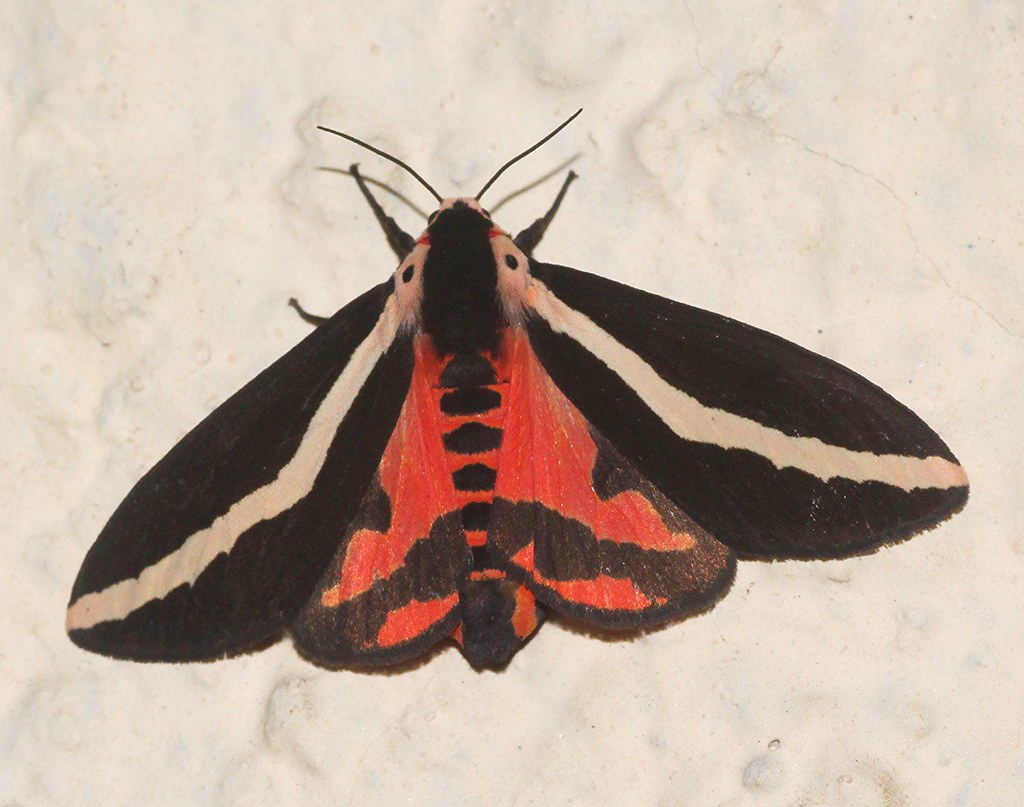}
  \caption{Figured Tiger Moth (\textit{Apantesis figurata})}
  \label{fig:1}
\end{subfigure}\hfil 
\begin{subfigure}{0.25\textwidth}
  \includegraphics[width=\linewidth]{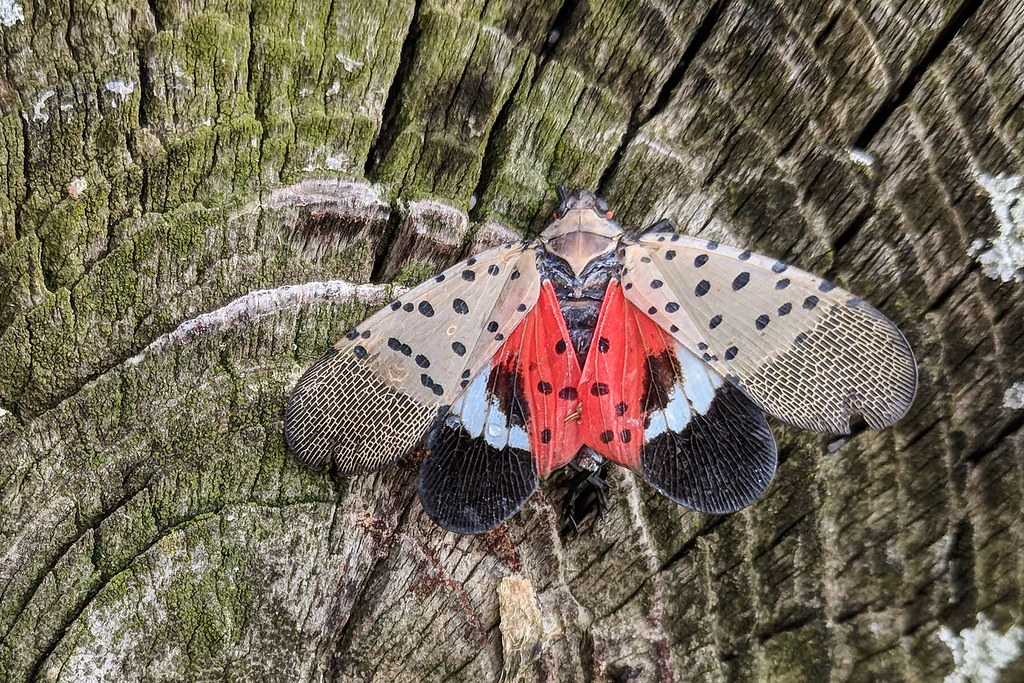}
  \caption{Spotted Lanternfly (\textit{Lycorma delicatula})}
  \label{fig:2}
\end{subfigure}\hfil 
\begin{subfigure}{0.25\textwidth}
  \includegraphics[width=\linewidth]{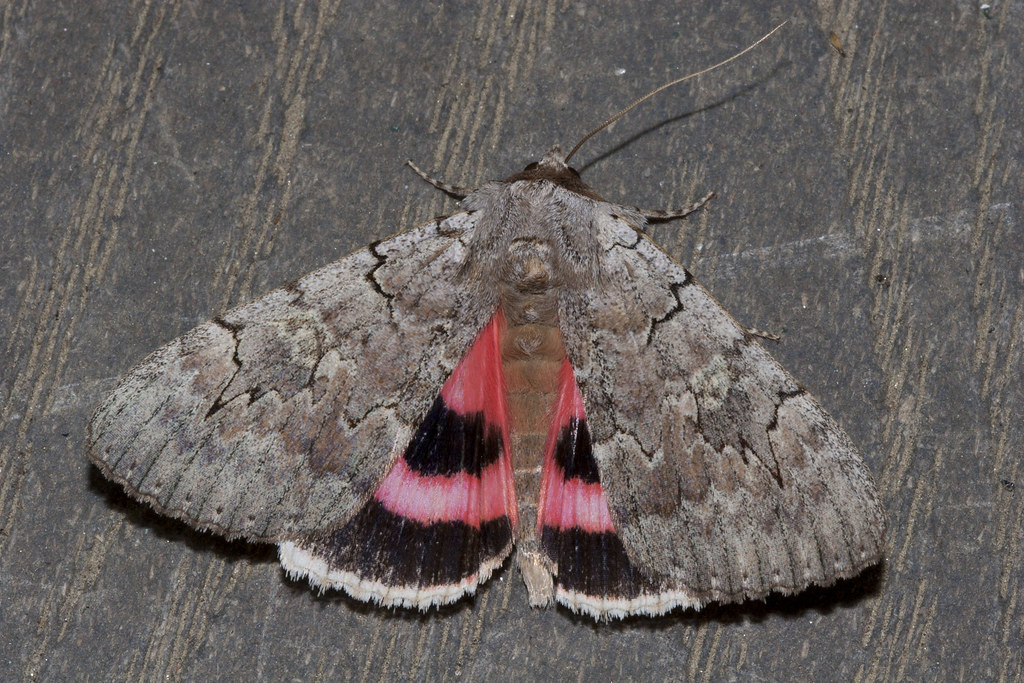}
  \caption{Pink Underwing (\textit{Phyllodes imperialis})}
  \label{fig:3}
\end{subfigure}

\medskip
\begin{subfigure}{0.25\textwidth}
  \includegraphics[width=\linewidth]{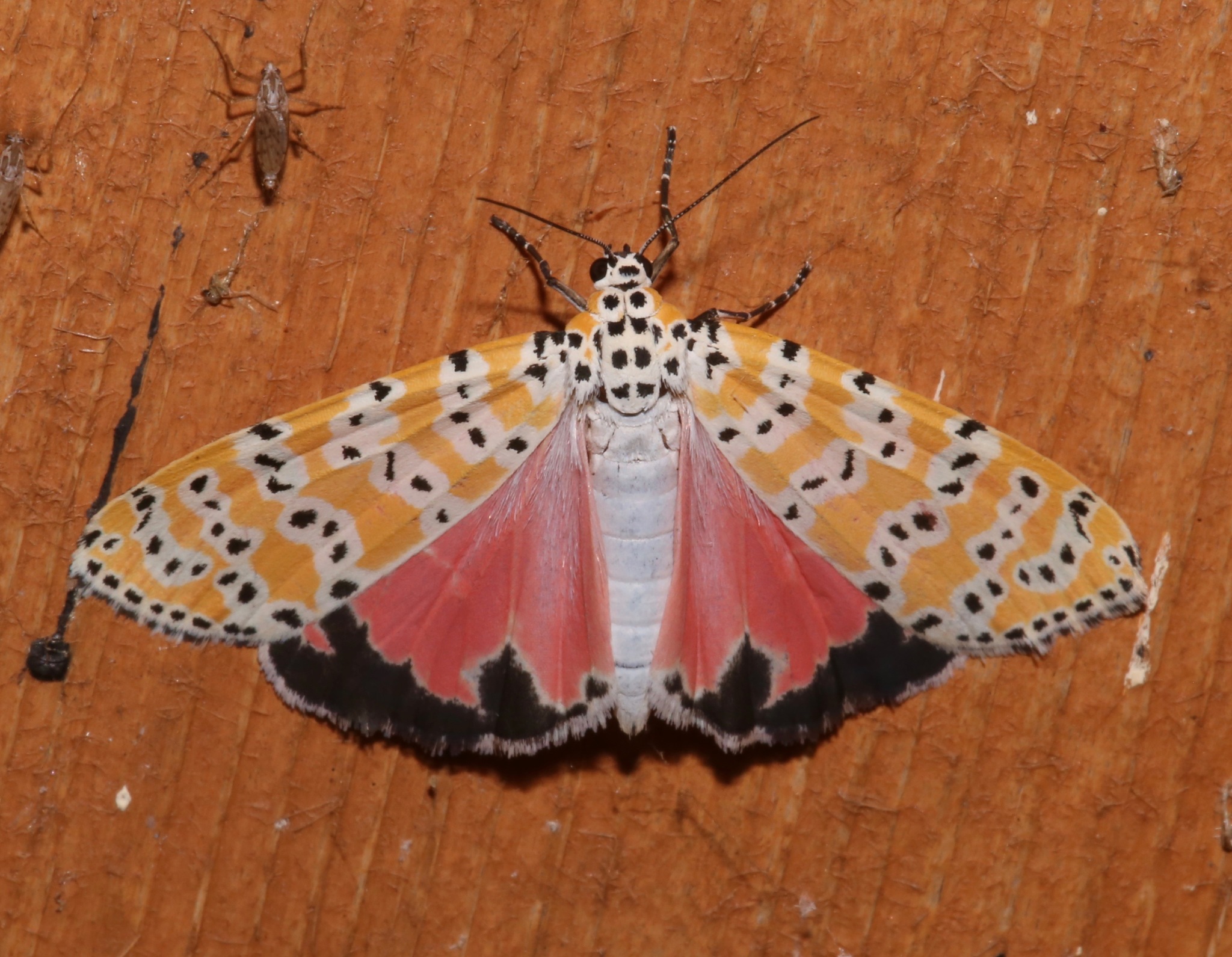}
  \caption{Bella Moth (\textit{Utetheisa ornatrix})}
  \label{fig:4}
\end{subfigure}\hfil 
\begin{subfigure}{0.25\textwidth}
  \includegraphics[width=\linewidth]{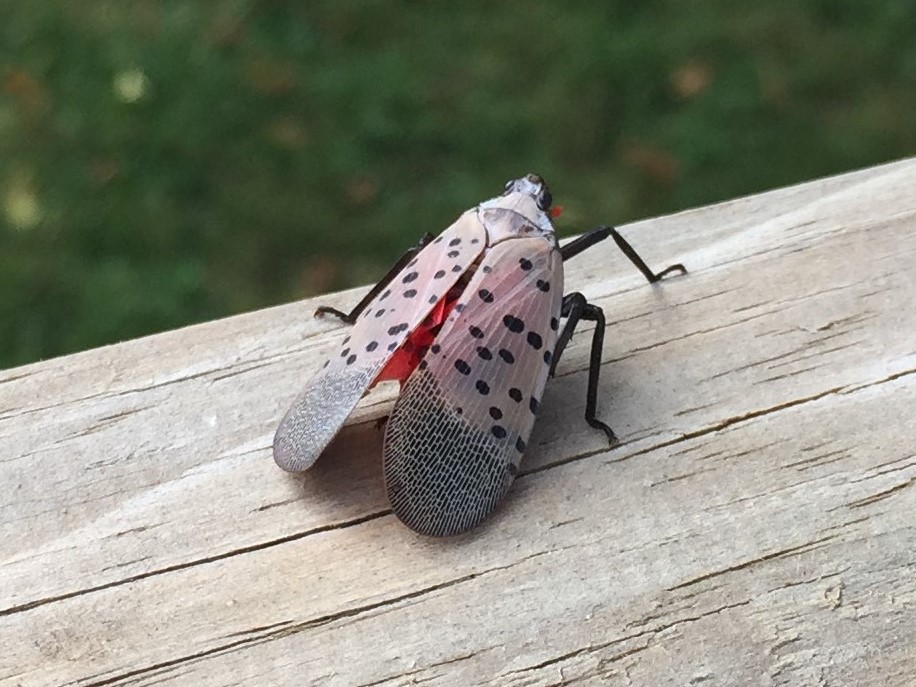}
  \caption{Spotted Lanternfly, wings closed}
  \label{fig:5}
\end{subfigure}\hfil 
\begin{subfigure}{0.25\textwidth}
  \includegraphics[width=\linewidth]{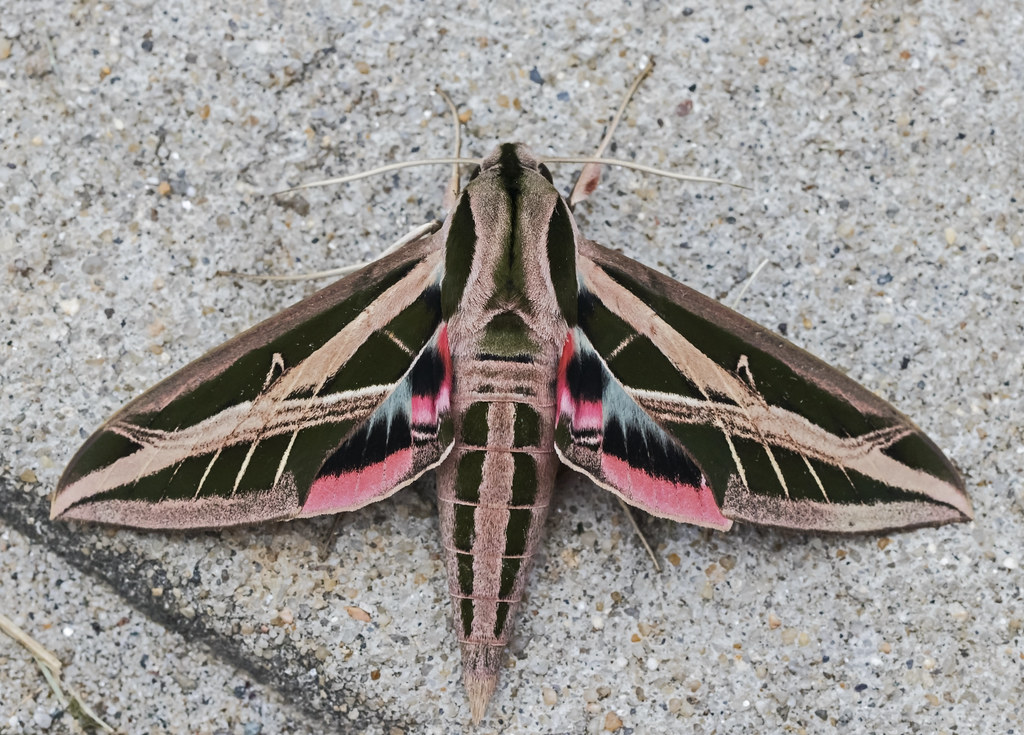}
  \caption{White-lined Sphinx Moth (\textit{Hyles lineata})}
  \label{fig:6}
\end{subfigure}
\caption{The SLF and its look-alikes. Similar body-shape, color, and the presence of a bright pink, orange, or red underwing contribute to the visual similarity of these insects.}
\label{fig:images}
\end{figure*}

The expanding scope of the invasive SLF problem provides ground for computer vision applications to enhance research studies and containment efforts. For example, SLF pose estimation may boost understanding of key population-growth activities such as egg-laying and enable motion tracking to characterize population spread \cite{pose1, pose2}. Additionally, while ecologists studying the spatial distribution of the SLF have long relied on manual surveying methods in order to collect data \cite{spatial1, spatial2, 1, 11}, advancements in AI animal detection techniques \cite{detection1, detection2} may improve the efficiency of such data collection efforts. Finally, the presence of insect species that are visually similar to the SLF (\cref{fig:images}) raises additional complications for both researchers and the public, as misidentification of such insects affects the quality of collected data and results in unnecessary harm to native wildlife. For this issue, there is significant potential for AI-based classifiers to assist ecologists and promote public management efforts. To enable such advancements in computer vision research for the invasive spotted lanternfly problem, there is a great need for a high-quality public dataset of SLF images.

To solve this problem, we present in this work a framework that consists of: (i) \textbf{LANTERN-RD}, a novel image dataset of the invasive SLF and visually-similar insects in order to enable computer vision research for SLF mitigation, (ii) a corresponding \textbf{baseline convolutional neural network} (CNN), and (iii) a \textbf{mobile deployment of the classifier} to empower rapid identification of the SLF against look-alikes and advocate for responsible mitigation efforts.

\section{Dataset}
\label{sec:dataset}

For the purpose of empowering the development of deep learning models to assist in efforts to contain the invasive SLF, we assemble an image dataset of the spotted lanternfly and its look-alikes, as summarized in \cref{fig:images}. 
 
\subsection{Data Overview}
\begin{figure}[t]
  \centering
   \includegraphics[width=0.8\linewidth]{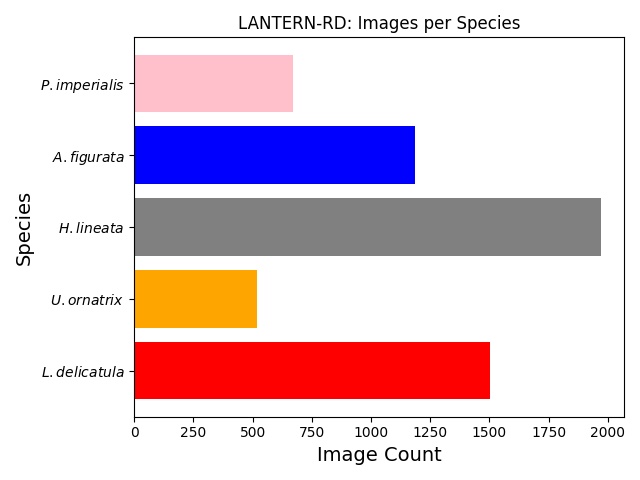}

   \caption{Distribution of Images Per Species.}
   \label{fig:distribution}
\end{figure}

The first iteration of LANTERN-RD includes a total of 5 insect classes: \textit{L. delicatula}, \textit{A. figurata}, \textit{P. imperialis}, \textit{U. ornatrix}, and \textit{H. lineata}, with 1187, 1501, 672, 520, and 1970 images, respectively. The distribution is visualized in \cref{fig:distribution}, and the total size of the dataset is 5850 images, with each image containing exactly one insect. Sample images of the dataset are displayed in \cref{fig:images}. The data are diverse, including images of insects in different poses, assorted backgrounds, and varied lighting conditions, allowing for the training of generalized models and the generation of sub-datasets for specific tasks. The dataset is presented as several files of image URLs, grouped by class, alongside a corresponding document outlining labels. 

\subsection{Pipeline}

In order to curate a high-quality dataset, we employ a robust pipeline to gather, clean, and compile data. 

\textbf{Extraction.} Raw image data are sourced from Bing Images and LAION \cite{22}. A number of strategic queries are utilized to capture a large volume of image URLs, which are stored separately according to class. However, the collected data are noisy and must be filtered before proceeding. 

\textbf{Automated Filtration.} In the first step of cleaning, we use several automated techniques to eliminate extraneous data. We begin with image hashing to remove duplicate and near-duplicate images. Next, when image captions are available (as with the LAION data), we parse these captions and eliminate data containing phrases linked to insects other than the queried class. Through this process, we are able to remove a large volume of noisy data. 

\textbf{Additional Filtration.} We parse the remaining data via several methods, removing the remaining extraneous images. For example, ``Tiger Moth" also happens to be the name of a popular 20th century biplane, and images of this aircraft were highly represented in the raw data before additional cleaning. We also carefully inspect each image and apply established identification methods to verify that the insect pictured belongs to the label.

The filtration process leaves a collection of image URLs that constitute the final, cleaned dataset. This modular pipeline allows for the efficient integration of new data sources such as community image submissions, enables expansion to additional classes of look-alike species, and takes strides to ensure that the data are clean and useful for training. 

\section{Baseline Experiments}
\label{sec:CNN}

Alongside the dataset, we train a baseline CNN using the VGG16 \cite{vgg} model architecture. The data are randomly split for training, validation, and testing, with a ratio of 60\%, 20\%, and 20\%, respectively. Before training, the image data are augmented. Each class pictured in \cref{fig:images} receives a numerical label, and the CNN is trained according to the categorical cross entropy loss function for 50 epochs at a rate of 1e-3.

\subsection{Results}
The model achieved an overall test accuracy of 97.20\%. The per-class F1 scores are summarized in \cref{tab:f1}.

\begin{table}
  \centering
  \begin{tabular}{@{}lc@{}}
    \toprule
    Species & F1 Score \\
    \midrule
    \textit{L. delicatula} (Spotted Lanternfly) & 0.983 \\
    \textit{A. figurata} (Tiger Moth) & 0.946 \\
    \textit{P. imperialis} (Pink Underwing) & 0.935\\
    \textit{U. ornatrix} (Bella Moth) & 0.984 \\
    \textit{H. lineata} (White-Lined Sphinx Moth) & 0.989 \\
    \bottomrule
  \end{tabular}
  \caption{The preliminary CNN achieves the above results per-class.}
  \label{tab:f1}
\end{table}

\section{Mobile Implementation}
\label{sec:mobile}

We additionally propose a simple mobile application that implements the classifier trained in \cref{sec:CNN}. Sensationalization of the spotted lanternfly in affected areas has resulted in the public becoming more closely involved in management efforts, with programs such as New Jersey's ``Stomp It Out!" \cite{njslf2} calling on citizens to exterminate sighted SLFs. As a result, it is imperative that the public is able to accurately identify this invasive insect against look-alike species, particularly to prevent well-intentioned citizen scientists from unnecessarily harming wildlife. 

\textbf{Features.} The app allows users to directly capture or upload an image of an insect suspected to be an SLF. Subsequently, the picture is fed as an input to the classifier, and the user is notified of the class prediction. All user data are kept private and operations are run locally.

This app presents one useful application of LANTERN-RD in training deep learning models to assist in efforts to contain the invasive SLF. We hope that such an application motivates further advancements from the computer vision community. 

\section{Conclusion}
\label{sec:conclusion}
In this paper, we have introduced LANTERN-RD, a curated dataset consisting of diverse images of the invasive spotted lanternfly and visually similar insects. This dataset contains 5850 images of the spotted lanternfly and four visually similar insects, and is curated via an efficient pipeline that is scalable to additional data sources and new classes. A baseline classifier trained on LANTERN-RD achieves a 97.20\% test accuracy. This validates that datasets such as LANTERN-RD will enable a wide array of computer vision applications that have positive downstream impacts on efforts to contain the invasive spotted lanternfly. To explore one avenue of computer vision applications for the invasive SLF problem, we implement the preliminary classifier into a simple app designed for users to use their mobile devices to rapidly understand whether or not an insect is the SLF. This assists in research activities and boosts caution on behalf of citizen scientists. Future work would include solicitation of community image submissions in order to expand the scale of the dataset, particularly in new and underserved insect classes, and for different stages of the SLF lifecycle. We call on ecologists and the computer vision community to come together in applying LANTERN-RD for deep learning tasks poised to increase the efficiency, reliability, and scale of efforts to contain the invasive SLF.


{\small
\bibliographystyle{ieee_fullname}
\bibliography{egbib}
}

\end{document}